\def\year{2019}\relax
\documentclass[letterpaper]{article}
\usepackage{aaai19}
\usepackage{times}
\usepackage{helvet}
\usepackage{courier}
\usepackage{url}
\usepackage{graphicx}
\usepackage{algorithm}
\usepackage{algpseudocode}
\usepackage{amsmath}
\usepackage{amssymb}
\usepackage{mathtools}
\usepackage{tabu}

\newcommand{\routine}{\textit{routine}}
\newcommand{\queue}{\textit{queue}}
\newcommand{\start}{\textit{start}}
\newcommand{\result}{\textit{result}}

\title{Learning Classical Planning Strategies with Policy Gradient}
\author{
Pawe\l{} Gomoluch \and Dalal Alrajeh \and Alessandra Russo\\
Department of Computing, Imperial College London\\
\{pawel.gomoluch14,dalal.alrajeh,a.russo\}@imperial.ac.uk\\
}
\begin{document}
\maketitle
\begin{abstract}
A common paradigm in classical planning is heuristic forward search. Forward search planners often rely on simple best-first search which remains fixed throughout the search process. In this paper, we introduce a novel search framework capable of alternating between several forward search approaches while solving a particular planning problem. Selection of the approach is performed using a trainable stochastic policy, mapping the state of the search to a probability distribution over the approaches. This enables using policy gradient to learn search strategies tailored to a specific distributions of planning problems and a selected performance metric, e.g. the IPC score. We instantiate the framework by constructing a policy space consisting of five search approaches and a two-dimensional representation of the planner's state. Then, we train the system on randomly generated problems from five IPC domains using three different performance metrics. Our experimental results show that the learner is able to discover domain-specific search strategies, improving the planner's performance relative to the baselines of plain best-first search and a uniform policy.
\end{abstract}

\section{Introduction}
\label{sec:introduction}
As a simple and complete search algorithm, best-first search forms the core of many modern classical planners (e.g. \cite{Helmert2006,Richter2010}). Approaches combining greedy best-first search (GBFS) with other planning techniques have largely been confined to sequentially attempting to solve the problem using two different search modes (e.g. \cite{Hoffmann2001,Lipovetzky2017} or even an entire portfolio of potentially unrelated algorithms (e.g. \cite{Howe1999,Gerevini2009,Fewcett2011,Helmert2011,Cenamor2016}). Elsewhere, best-first search (BFS) has been combined with an auxiliary exploratory technique, triggered when the main GBFS fails to reach progress for a certain number of expansions \cite{Xie2014,Lipovetzky2017}.

In this work, we introduce a framework capable of systematically alternating between various forward search techniques, in the course of solving the planning problem. Unlike \cite{Xie2014,Lipovetzky2017}, we equip the planner with more than two techniques and do not manually specify the rules for choosing between them. The choice of the technique is made by the planner using a stochastic strategy, trained to maximize the planner's performance using reinforcement learning. With this approach, it is possible to train the planner for a particular domain or a selected performance objective, such as maximizing the IPC score or minimizing the time required to find a solution. By enabling the learner to discover domain-specific search strategies, this approach has the potential to cover the middle ground between general-purpose search algorithms designed to fit a variety of domains and the domain-specific solvers, hand-crafted by human experts. Moreover, through seamless integration of various performance objectives, it allows for automatic navigation of the trade-off between the time required to find plans and their cost, which is a central issue in satisficing planning.

To demonstrate applicability of our framework, we introduce a specific instantiation, which uses five different forward search approaches and a simple characterization of the planner's state in terms of the estimated distance to the goal (i.e. the heuristic value) and the remaining time available to the planner.

We then empirically evaluate the resulting system by training it on randomly generated problems from five IPC domains. Although our training scheme uses planning problems of relatively small size, we extend our evaluation to include larger, IPC-scale problems, which allows for testing the generality of the learned strategies with respect to the problem size.

\section{Related work}
\label{sec:related-work}

In \cite{Xie2014}, GBFS is augmented with local search or random walks whenever the search does not yield progress for a certain number of node expansions. Progress of the search is determined by decrease of $h_{min}$, the lowest value of the heuristic function observed so far. In a similar way, GBFS can be combined with \emph{width-based} search, which prunes out states which do not satisfy a novelty criterion \cite{Lipovetzky2017}. Both approaches rely on single addition to GBFS, which is triggered when the main search fails to find states with lower heuristic value for a certain number of expansions. Our framework allows for a larger number of search approaches and does not explicitly distinguish between primary and backup ones. The strategy for choosing between them is subject to the learning process.

Related to our approach is also the concept of portfolio planners, especially those configured based on experience gathered on a set of training problems \cite{Fewcett2011,Helmert2011,Cenamor2016}. 
The key difference between portfolio approaches and our work is that the operation of a portfolio planner involves running a number of independent searches, each with a different planning algorithm. In our approach a single search is performed, with the possibility of alternating between compatible search techniques, depending on the state of the search.

Learning from experience has long been used as a way of improving the planner's performance. In classical planning, the work in this area included learning macro actions (e.g. \cite{Fikes1972,Coles2007}, control knowledge in the form of decision rules (e.g. \cite{Leckie1998,Yoon2008}) and heuristic functions \cite{Yoon2008,Virseda2013,Garrett2016}. A survey of learning methods for automated planning can be found in \cite{Jimenez2012}. To the best of our knowledge, none of the learning approaches attempted to construct a domain-specific composition of different search techniques.

\section{Background}
\label{sec:background}
\paragraph{Classical planning}
Planning is the problem of finding sequences of actions which, when executed from a given initial state, lead to a state in which the planning goal is satisfied. Classical planning, in particular, relies on a known and perfect model of the environment, including a discrete set of deterministic actions. Formally, a classical planning task is given by a tuple $\langle V_p,O,s_o,g\rangle$, where $V_p$ is a set of finite-domain variables, $O$ is the set of operators, $s_0$ is the initial state, which is an assignment over the variables of $V_p$, and $g$ is the goal, a partial assignment over the variables of $V_p$. Each operator $o \in O$ is itself specified with a tuple $\langle \operatorname{pre}(o), \operatorname{eff}(o)\rangle$, where $\operatorname{pre}(o)$ and $\operatorname{eff}(o)$ are both partial assignments over $V_p$, defining the preconditions and the effects of applying operator $o$, respectively. Operator $o$ can be applied in state $s$ if and only if $\operatorname{pre}(o) \subseteq s$. The result of applying operator $o$ in state $s$, denoted as $o(s)$ is defined as an assignment over $V$ differing from $s$ by setting the variables covered by $\operatorname{eff}(o)$ accordingly. 

A common classical planning approach is forward search. The search starts at the initial state and iteratively explores states reachable by sequentially applying operators of set $O$. A canonical example of forward search is \emph{best-first search} (BFS), which always expands the node $n$ with the lowest value under a specified evaluating function $f(n)$. Heuristic functions guide forward search by estimating the distance to the goal from any given node. Best-first search driven solely by the heuristic value of a node, $f(n)=h(n)$, is known as \emph{greedy best-first search} (GBFS).

\paragraph{Reinforcement learning}
We adopt a reinforcement learning approach based on the standard Markov Decision Process (MDP) setting. An MDP is given by a tuple $\langle S, A, p(s,a,s'), r(s,a), \gamma\rangle$, where S is a set of states, A is a set of actions, $p(s,a,s'):S\times A\times S\rightarrow S$ is a function determining the probability of a transition to state $s'$, given that action $a$ is taken in state $s$, $r(s,a): S\times A \rightarrow \mathbb{R}$ is a function yielding the expected reward for taking action $a$ in state $s$ and $\gamma$ is the discount factor. The task of an agent operating in MDP is to maximize the discounted sum of rewards observed during an episode: $G = \sum_{t=1}^T\gamma^tr_t$, where $T$ is the length of the episode. A \emph{policy} of the agent is a (possibly stochastic) mapping from $S$ to $A$. Since in this paper we focus on stochastic policies (search strategies), we assume a policy to be a function $\pi(s,a) = p(a|s)$, indicating the probability of agent selecting action $a$ in state $s$, for every $s \in S, a \in A$.

Given the focus on stochastic policies, we employ a policy gradient method of learning them, based on the REINFORCE algorithm \cite{Williams1992,Sutton1998}. The core idea of REINFORCE is that unbiased samples of the gradient of the return $G$ with respect to parameters $\theta$ of policy $\pi$ can be computed using an expression that only depends on the current policy and a single return sample. Every state-action pair $(s,a)$ occurring at time $t$ in a particular episode, generates the following update of the policy parameters $\theta$:
\begin{equation}
\Delta\theta = \alpha (G_t - b(s))\frac{\nabla_{\theta}\pi(s,a)}{\pi(s,a)}
\end{equation}
where $\alpha$ is the learning rate, $G_t$ is the discounted sum of rewards from time $t+1$ to the end of the episode and $b(s)$ is the baseline for state $s$, for example computed as the average of returns observed for that state in previous episodes.

\section{Approach}
\label{sec:approach}
The planning approach proposed in this paper relies on the idea that the search algorithm does not have to be fixed throughout the process of solving a planning problem. A range of planning algorithms can interleave, provided that they can all be cast as operations processing common internal state of the planner. In the remainder of the paper, we assume that the algorithms are available to the planner in the form of a set of \emph{routines}. The routines applied to the state of the planner perform an atomic step of the corresponding algorithm. For example, a step of GBFS can consist of selecting the node with the lowest heuristic value, expanding it and adding its children to the queue (open list).

Algorithm \ref{planning-asr} outlines our framework for planning with alternating search routines. Typically for a planner, the algorithm's arguments include the initial state $s_0$, the goal $g$ and the set of operators $O$. Additional parameters are the set of search routines $A$ and time limit $t_r$. The algorithm initializes the state \emph{queue} (the open list) with $s_0. $Then, it chooses a routine from set of routines $A$, applies it for a limited time of up to $t_r$, then chooses a routine again, and so on, until a plan is found. A single application of \routine\ to the \emph{queue} modifies the queue and returns the plan if a solution is found, returns failure if the search space is exhausted without finding a plan and returns a special \emph{in-progress} token otherwise. Note that the algorithm is presented in a simplified form, focusing on alternating between the routines. It hides details such as keeping track of already expanded states (closed list) and recording their ancestors for the purpose of extracting the plan.

\begin{algorithm}
\caption{Planning with alternating search routines} \label{planning-asr}
\begin{algorithmic}
\Function{PlanASR}{$s_0, g, O, A, t_r$}
\State $\queue \gets [s_0]$
\While{true}
\State $\routine \gets \operatorname{choose}(A)$
\State $t_{\textrm{\start}} \gets \operatorname{now}() $ \Comment{current time}
\While{$ \operatorname{now}()-t_{\textrm{\start}} < t_r $}
\State $\result \gets \routine(\queue)$
\If{$\result \neq \operatorname{in-progress}$}
\State $\operatorname{return}\result$ \Comment{a plan or failure}
\EndIf
\EndWhile
\EndWhile
\EndFunction
\end{algorithmic}
\end{algorithm}

The problem of choosing a routine given the state of the search can be modeled as an MDP where the state set $S$ is the set of possible states of the search and the action set $A$ is the set of search routines available to the planner. The states of the search are assignments over a set of finite-domain variables $V_l$. In Section \ref{sec:implementation} we introduce particular instantiations of $A$ and $V_l$. The reward function can be any measure of planner's performance available after a single attempt at a given planning problem. In the simplest scenario, the reward can be defined as 1 if the planner solves the problem within a set time limit and 0 otherwise. Training with such a reward function would correspond to optimizing for coverage, i.e. the number of problems solved, disregarding the plan quality. The reward functions used in our experimental setup are discussed in Section \ref{sec:implementation}.

The search policy is parametrized by $\theta$, with one parameter $\theta_{i,j}$ for every state-action pair $(s_i, a_j)$. The probability of taking action $a_j$ in state $s_i$ is determined by softmax over the parameters associated with the state:
$$ \pi(a_{j}|s_{i}) = \frac{e^{\theta_{i,j}}}{\displaystyle\sum_{j}{e^{\theta_{i,j}}}} $$

We train the weights using a variant of episodic REINFORCE \cite{Williams1992}.
To decrease the variance of the gradient sample we average the update over $N$ episodes, during which the same stochastic policy is used on the same training problem. In our setting with just four states, a state is typically visited many times during a single episode and many (possibly different) actions are taken from it. The parameter update after every $N$ episodes is therefore:
\begin{equation}
\label{eq:update-rule}
\Delta\theta = \alpha\frac{1}{N}\displaystyle\sum_{i=1}^{N}\displaystyle\sum_{s,a}\eta_i(s, a)(G_t-V(s))\frac{\nabla_\theta\pi(s,a)}{\pi(s,a)}
\end{equation}
where $\alpha$ is the learning rate, $\eta_i(s,a)$ is the number of times action $a$ was taken from state $s$ in episode $i$, $G_t$ is the return observed at the end of the episode and $V(s)$ is a baseline for state $s$, computed as the average of all past returns for episodes passing through $s$.

\section{Approach instantiation}
\label{sec:implementation}

In this section we describe an instantiation of the framework described in Section \ref{sec:approach}, which we then empirically evaluate in Section \ref{sec:evaluation}. The key components of the framework are the set of search routines $A$, the representation of the planner's state and the reward function, reflecting the chosen performance objective.

Another important choice is $t_r$, the time span over which the chosen routine continues to be applied. In this work we fix $t_r$ to 100 ms, which enables a single routine to make substantial progress, while also allowing for strategies interleaving the routines in a fine-grained manner.

\subsection{Search routines}
Below, we describe the routines which we include in the set $A$ of our instantiation and state how they can be integrated within the framework.

\paragraph{Greedy best-first search} Plain greedy best-first search always expands the node with the lowest value of $h$. A single application of this routine consists of a single node expansion, followed by placing of all its ancestors in the queue.

\paragraph{$\epsilon$-greedy search} $\epsilon$-greedy search was first considered in classical planning context by \cite{Valenzano2014}. Like greedy best-first search, one application of this routine performs a single node expansion. The difference is that with probability of $\epsilon$ a random node is selected from the queue with the probability of selection uniform across all the nodes. Throughout the paper we use $\epsilon=0.2$.

\paragraph{Greedy search with random walks} A variation of GBFS, following the expansion of node $n$ with a single random walk of length $l$ starting in $n$, provided that no decrease in heuristic value has been observed for the last $s$ node expansions. All the nodes along the walk are added to the global queue. The walk stops as soon as a state with heuristic value lower than that of $n$ is found. This method is inspired by \cite{Xie2014}, but throughout the paper we use parameters of $s=5$ and $l=20$, which makes the random walks much more frequent. This is to ensure that the routine is substantially different from plain GBFS and offers the learner a meaningful alternative.

\paragraph{Local search} Local search is started from a node with the lowest $h$ value, extracted from the global queue when the routine is selected. The node is used to initialize the local queue, which persists between subsequent calls of the routine. The search continues by expanding states from the local queue. When the time limit $t_r$ expires, the local queue is merged into the global one (all the states from local queue are inserted to the global one). If the local queue becomes empty before $t_r$ expires, another node from the global queue is put in the local one, effectively starting a new local search.

\paragraph{Heuristic-guided depth-first search} Depth-first search is performed using a local search stack. When the routine is selected, a node with the lowest $h$ value is extracted from the global queue and placed on the stack. At every call, a node $n$ is popped from the stack and expanded. Descendant nodes are put on the stack in order of decreasing $h$ value, so that the node expanded at the next step is the descendant of $n$ with the lowest $h$. The descendant nodes are also inserted in the global queue. If the stack becomes empty before $t_r$ expires, another node from the global queue is put on the stack and the search continues.

\subsection{State representation}
As stated in Section \ref{sec:approach}, we the state space of our learner is the set of possible assignments over the variables of $V_l$. In the remainder of the paper, we consider $V_l$ to be a set of two boolean variables $V_l=\{d,t\}$. We take $d$ to be a binarized heuristic estimate of the distance to the goal:
$$ d =\begin{cases}
0& \text{if $ h_{\textrm{best}} < h_0 / 2 $}\\
1& \text{otherwise}
\end{cases}$$
where $h_{\textrm{best}}$ is the lowest heuristic value recorded in the search so far and $h_0$ is the heuristic value of the initial state. The variable is a simple way of tracking the progress of the search. Similarly, $t$ indicates how much time the planner has left:
$$ t =\begin{cases}
0& \text{if $ t_{\textrm{elapsed}} < t_{\textrm{max}} / 2$}\\
1& \text{otherwise}
\end{cases}
$$
where $t_{\textrm{elapsed}}$ is the time elapsed from the beginning of the search and $t_{\textrm{max}}$ is the total time allocated for the search.

Using two binary state features allows for a compact tabular representation of the learned policies. For readability, we further refer to the states of the search as \emph{near} ($d=0$) or \emph{far} ($d=1$) and \emph{early} ($t=0$) or \emph{late} ($t=1$).

\subsection{Reward functions}
We consider three different reward functions, corresponding to three different ways of scoring performance of a planner. The first reward is based on IPC score, first used in IPC-2008\footnote{\url{http://icaps-conference.org/ipc2008/deterministic/}}. This is a widely used measure of planners' performance, assigning higher scores to planners finding lower-cost solutions. For every solved problem, the planner receives score defined as follows:
$$ z = \frac{c_{min}}{c} $$
where $c$ is the cost of the plan returned by the planner and $c_{min}$ is the cost of best known solution. For failed problems, planners receive a score of 0.
Cost $c_{min}$ is determined using a set of reference planners, each based on a single routine from set $A$ and run with the same timeout as the trained planner. 

During the training phase, we deviate from traditional IPC score in two ways. First, we do not include the trained planner in the reference set, allowing for a situation where $c < c_{min}$ an so $z > 1$. This way, the learner observes rewards higher than 1 for finding solutions with costs strictly lower than any of the reference planners. Second, if the trained planner solves the problem but none of the reference planners does, we set the reward to a fixed value of 2. The choice of value 2 is motivated by the fact that, during a competition, solving a problem unsolved by others yields a net advantage of 1, which is 1 more than in the case when a solution with cost equal to the best competitor is found, when both planners receive one point.  Both of these changes aim at providing the reward the ability to distinguish between situations where the learner's performance is as good as the reference planners' and the cases when it is strictly better. If a problem is solved neither by the learner nor by any of the reference planners, the reward is not defined and the episode does not generate a parameter update. Formally,
$$ r_{ipc} =\begin{cases}
\frac{c_{min}}{c_L}& \text{if both $c_{min}$ and $c_L$ are defined}\\
0& \text{if only $c_{min}$ is defined}\\
2& \text{if only $c_L$ is defined}
\end{cases}
$$
where $c_L$ is the cost of the plan computed by the trained strategy.

The modified version of IPC score is only used in the training phase. Naturally, during the evaluation of the trained policies, the learned planner is included in the reference set, treating it on par with any other planner included in the comparison and effectively capping its score for any single problem at 1.

The second reward function we consider is based on squared IPC score. The motivation is to further increase emphasis on the cost of the plan. For example, a planner returning a plan two times more expensive than the best plan known would only receive $(\frac{1}{2})^2=\frac{1}{4}$ points. Given the modifications we introduced in $r_{ipc}$, we cap the reward at 2 to prevent excessive premium for performance strictly better than that of reference planners. Formally:
$$ r_{ipc2} = \operatorname{min}(r_{ipc}^2, 2) $$

Finally, we consider a reward based solely on the time used to find a solution. The learner receives a reward equal to the proportion of spared time to the total time allocated for the problem:
$$ r_{time} =\begin{cases}
\frac{T-t}{T}& \text{if solved} \\
0& \text{otherwise}
\end{cases}
$$
where $T$ is the total time allocated for the planner and $t$ is the actual time elapsed before the plan is returned. Note that another reward function disregarding the plan cost could be derived from coverage (1 if the problem is solved and 0 otherwise). This however would ignore the computation time as long as it falls within the limit. The time-based formulation retains more information in the reward signal and incentivizes finding a solution as quickly as possible. 

\section{Evaluation}
\label{sec:evaluation}

 \begin{table*}
\centering
{\small
\begin{tabu}{X[c,m]X[c,m]}
\textbf{Transport}\vspace{1ex} & \textbf{Parking}\vspace{1ex} \\
	\begin{tabu}{|c|ccccc|}
    \hline
    \multicolumn{6}{|c|}{IPC reward}\\\hline
    & GBFS & $\epsilon$-greedy & RW & Local & DFS \\\hline
    near early & \textbf{0.98} & 0.01 & $\sim$ 0 & $\sim$ 0 & $\sim$ 0 \\
    near late & 0.01 & $\sim$ 0 & \textbf{0.97} & 0.02 & $\sim$ 0 \\
    far early & $\sim$ 0 & $\sim$ 0 & $\sim$ 0 & \textbf{1.00} & $\sim$ 0 \\
    far late & 0.01 & 0.01 & \textbf{0.94} & 0.01 & 0.02 \\\hline
    \multicolumn{6}{|c|}{IPC\^{}2 reward}\\\hline
    & GBFS & $\epsilon$-greedy & RW & Local & DFS \\\hline
    near early & 0.01 & \textbf{0.98} & $\sim$ 0 & $\sim$ 0 & $\sim$ 0 \\
    near late & 0.14 & \textbf{0.22} & \textbf{0.60} & 0.04 & $\sim$ 0 \\
    far early & \textbf{0.96} & 0.02 & 0.01 & $\sim$ 0 & $\sim$ 0 \\
    far late & 0.02 & 0.02 & \textbf{0.91} & 0.05 & 0.01 \\\hline
    \multicolumn{6}{|c|}{Time reward}\\\hline
    & GBFS & $\epsilon$-greedy & RW & Local & DFS \\\hline
    near early & 0.01 & 0.01 & \textbf{0.95} & 0.01 & 0.01 \\
    near late & 0.04 & 0.03 & 0.05 & 0.09 & \textbf{0.78} \\
    far early & 0.01 & $\sim$ 0 & \textbf{0.97} & 0.01 & 0.01 \\
    far late & 0.19 & 0.17 & \textbf{0.23} & 0.19 & \textbf{0.21} \\\hline
    \end{tabu}
    &
    \begin{tabu}{|c|ccccc|}
    \hline
    \multicolumn{6}{|c|}{IPC reward}\\\hline
    & GBFS & $\epsilon$-greedy & RW & Local & DFS \\\hline
    near early & $\sim$ 0 & $\sim$ 0 & $\sim$ 0 & $\sim$ 0 & \textbf{0.99} \\
    near late & 0.01 & 0.01 & 0.01 & 0.02 & \textbf{0.94} \\
    far early & $\sim$ 0 & $\sim$ 0 & $\sim$ 0 & $\sim$ 0 & \textbf{0.99} \\
    far late & 0.2 & 0.2 & 0.2 & 0.2 & 0.2 \\\hline
    \multicolumn{6}{|c|}{IPC\^{}2 reward}\\\hline
    & GBFS & $\epsilon$-greedy & RW & Local & DFS \\\hline
    near early & 0.2 & 0.02 & 0.01 & 0.15 & \textbf{0.80} \\
    near late & 0.01 & 0.01 & 0.01 & 0.02 & \textbf{0.96} \\
    far early & $\sim$ 0 & $\sim$ 0 & 0.01 & $\sim$ 0 & \textbf{0.98} \\
    far late & 0.19 & 0.19 & 0.19 & 0.19 & \textbf{0.25} \\\hline
    \multicolumn{6}{|c|}{Time reward}\\\hline
    & GBFS & $\epsilon$-greedy & RW & Local & DFS \\\hline
    near early & $\sim$ 0 & $\sim$ 0 & $\sim$ 0 & 0.01 & \textbf{0.98} \\
    near late & 0.19 & 0.19 & 0.18 & 0.19 & \textbf{0.25} \\
    far early & 0.01 & 0.01 & 0.01 & 0.01 & \textbf{0.97} \\
    far late & 0.2 & 0.2 & 0.2 & 0.2 & 0.2 \\\hline
    \end{tabu}
    \\
    \\
\textbf{Elevators}\vspace{1ex} & \textbf{No-mystery}\vspace{1ex}\\
    \begin{tabu}{|c|ccccc|}
    \hline
    \multicolumn{6}{|c|}{IPC reward}\\\hline
    & GBFS & $\epsilon$-greedy & RW & Local & DFS \\\hline
    near early & 0.05 & \textbf{0.58} & 0.03 & \textbf{0.32} & 0.02 \\
    near late & 0.01 & $\sim$ 0 & 0.01 & \textbf{0.98} & $\sim$ 0 \\
    far early & 0.02 & 0.01 & \textbf{0.92} & 0.02 & 0.04 \\
    far late & 0.11 & 0.09 & 0.11 & 0.16 & \textbf{0.53} \\\hline
    \multicolumn{6}{|c|}{IPC\^{}2 reward}\\\hline
    & GBFS & $\epsilon$-greedy & RW & Local & DFS \\\hline
    near early & 0.01 & $\sim$ 0 & \textbf{0.98} & $\sim$ 0 & $\sim$ 0 \\
    near late & 0.01 & 0.01 & \textbf{0.97} & 0.01 & $\sim$ 0 \\
    far early & 0.01 & 0.03 & \textbf{0.45} & \textbf{0.51} & $\sim$ 0 \\
    far late & 0.14 & 0.14 & 0.19 & \textbf{0.28} & \textbf{0.25} \\\hline
    \multicolumn{6}{|c|}{Time reward}\\\hline
    & GBFS & $\epsilon$-greedy & RW & Local & DFS \\\hline
    near early & 0.02 & 0.01 & 0.09 & 0.08 & \textbf{0.79} \\
    near late & 0.07 & 0.04 & \textbf{0.30} & \textbf{0.38} & \textbf{0.21} \\
    far early & 0.03 & 0.03 & 0.04 & 0.04 & \textbf{0.86} \\
    far late & 0.19 & 0.19 & \textbf{0.21} & 0.20 & \textbf{0.21} \\\hline
    \end{tabu}
    &
    \begin{tabu}{|c|ccccc|}
    \hline
    \multicolumn{6}{|c|}{IPC reward}\\\hline
    & GBFS & $\epsilon$-greedy & RW & Local & DFS \\\hline
    near early & $\sim$ 0 & \textbf{0.93} & $\sim$ 0 & 0.06 & 0.01 \\
    near late & \textbf{0.94} & 0.05 & 0.01 & $\sim$ 0 & $\sim$ 0 \\
    far early & 0.01 & \textbf{0.86} & 0.01 & 0.01 & 0.10 \\
    far late & 0.18 & 0.19 & \textbf{0.24} & 0.20 & 0.19 \\\hline
    \multicolumn{6}{|c|}{IPC\^{}2 reward}\\\hline
    & GBFS & $\epsilon$-greedy & RW & Local & DFS \\\hline
    near early & 0.06 & \textbf{0.87} & 0.01 & 0.02 & 0.04 \\
    near late & 0.01 & \textbf{0.99} & $\sim$ 0 & $\sim$ 0 & $\sim$ 0 \\
    far early & 0.01 & \textbf{0.92} & 0.02 & 0.03 & 0.02 \\
    far late & 0.19 & \textbf{0.22} & 0.20 & 0.20 & 0.20 \\\hline
    \multicolumn{6}{|c|}{Time reward}\\\hline
    & GBFS & $\epsilon$-greedy & RW & Local & DFS \\\hline
    near early & 0.01 & \textbf{0.97} & 0.01 & $\sim$ 0 & 0.01 \\
    near late & \textbf{0.42} & \textbf{0.38} & 0.09 & 0.05 & 0.06 \\
    far early & 0.01 & \textbf{0.96} & 0.01 & 0.01 & 0.01 \\
    far late & 0.2 & 0.2 & 0.2 & 0.2 & 0.2 \\\hline
    \end{tabu}
    \\
    \\
	\textbf{Floortile}\vspace{1ex}\\
    \begin{tabu}{|c|ccccc|}
    \hline
    \multicolumn{6}{|c|}{IPC reward}\\\hline
    & GBFS & $\epsilon$-greedy & RW & Local & DFS \\\hline
    near early & $\sim$ 0 & \textbf{1.00} & $\sim$ 0 & $\sim$ 0 & $\sim$ 0 \\
    near late & \textbf{0.42} & \textbf{0.55} & $\sim$ 0 & 0.01 & $\sim$ 0 \\
    far early & 0.01 & \textbf{0.98} & $\sim$ 0 & $\sim$ 0 & $\sim$ 0 \\
    far late & 0.2 & 0.2 & 0.2 & 0.2 & 0.2 \\\hline
    \multicolumn{6}{|c|}{IPC\^{}2 reward}\\\hline
    & GBFS & $\epsilon$-greedy & RW & Local & DFS \\\hline
    near early & $\sim$ 0 & \textbf{1.0} & $\sim$ 0 & $\sim$ 0 & $\sim$ 0 \\
    near late & 0.02 & \textbf{0.97} & $\sim$ 0 & $\sim$ 0 & $\sim$ 0 \\
    far early & 0.01 & \textbf{0.97} & 0.01 & 0.01 & 0.01 \\
    far late & 0.2 & 0.2 & 0.2 & 0.2 & 0.2 \\\hline
    \multicolumn{6}{|c|}{Time reward}\\\hline
    & GBFS & $\epsilon$-greedy & RW & Local & DFS \\\hline
    near early & $\sim$ 0 & \textbf{1.00} & $\sim$ 0 & $\sim$ 0 & $\sim$ 0 \\
    near late & \textbf{0.79} & 0.19 & 0.01 & 0.01 & 0.01 \\
    far early & 0.01 & \textbf{0.97} & 0.01 & 0.01 & 0.01 \\
    far late & 0.2 & 0.2 & 0.2 & 0.2 & 0.2 \\\hline
    \end{tabu}&
    \parbox{0.3\textwidth}{Entries in the table are the probabilities of selecting a given routine in a given search state, learned separately for each of the domains and reward functions. Probabilities $> 0.2$ highlighted in \textbf{bold}. The impact of different rewards is most visible in \emph{Transport} and \emph{Elevators} domains, where time-based reward shifts the policies towards depth-first search. On the other hand, changing the reward has little effect in \emph{No-mystery} and \emph{Floortile} domains, where the more aggressive routines are potentially harmful, and in \emph{Parking}, which admits DFS without big impact on the plan cost.} 
\end{tabu}}
\caption{The policies learned for each of the planning domains and reward functions.}
\label{tab:policies}
\end{table*}

\begin{table}[ht]
\centering
{\small
\begin{tabu}{|c|ccccc|c|}
\hline
\multicolumn{7}{|c|}{IPC score}\\\hline
& T & P & E & N & F & Sum \\\hline
GBFS & 29.46 & 15.57 & 28.73 & 24 & 28.27 & 126.03\\
$\epsilon$-gr.& 27.23 & 10.88 & 21.87 & \textbf{30.64} & \textbf{50.33} & 140.95\\
RW & 30.92 & 9.79 & 31.82 & 19.48 & 4.91 & 96.92\\
Local & 37.18 & 21.89 & \textbf{32.87} & 20 & 23.9 & 135.84\\
DFS & 12.95 & \textbf{37.45} & 16.92 & 7.7 & 0.13 & 75.15\\
Uni & 24.07 & 25.71 & 26.79 & 20.75 & 32.44 & 129.76\\
L(I) & \textbf{38.85} & \textbf{37.27} & \textbf{32.67} & 26.62 & \textbf{49.83} & \textbf{185.24}\\
L(I\^{}2) & 36.79 & \textbf{37.38} & \textbf{32.13} & 27.2 & 49 & 182.5\\
L(T) & 30.58 & \textbf{36.56} & 19.54 & 26.37 & 46.84 & 159.89\\\hline
\multicolumn{7}{|c|}{IPC\^{}2 score}\\\hline
& T & P & E & N & F & Sum \\\hline
GBFS & 27.49 & 13.66 & \textbf{27.6} & 24 & 24.67 & 117.42\\
$\epsilon$-gr.& 22.82 & 9.52 & 20.87 & \textbf{30.5} & \textbf{44.69} & 128.4\\
RW & 19.26 & 8.41 & 24.27 & 19.38 & 3.95 & 75.27\\
Local & 28.58 & 16.76 & 26.44 & 20 & 20.54 & 112.32\\
DFS & 4.07 & \textbf{25.93} & 6.42 & 7.7 & 0.06 & 44.18\\
Uni & 13.73 & 19.71 & 18.24 & 20.71 & 28.33 & 100.72\\
L(I) & \textbf{30.7} & \textbf{26.18} & 25.13 & 26.55 & \textbf{44.28} & \textbf{152.84}\\
L(I\^{}2) & \textbf{30.55} & \textbf{26.51} & 24.57 & 27.1 & 43.59 & \textbf{152.32}\\
L(T) & 18.75 & 25.5 & 8.77 & 26.26 & 41.36 & 120.64\\\hline
\multicolumn{7}{|c|}{Time score}\\\hline
& T & P & E & N & F & Sum \\\hline
GBFS & 23.66 & 14.27 & 14.83 & \textbf{20.58} & 17.13 & 90.47\\
$\epsilon$-gr.& 21.48 & 10.15 & 10.44 & \textbf{21.58} & \textbf{38.47} & 102.12\\
RW & 39.93 & 9.35 & 21.28 & 16.77 & 3.96 & 91.29\\
Local & 33.87 & 22.45 & 19.22 & 15.98 & 14.02 & 105.54\\
DFS & 36.88 & \textbf{48.6} & \textbf{25.41} & 5.44 & 0.21 & 116.54\\
Uni & 34.55 & 27.75 & 20.25 & 17.18 & 19.71 & 119.44\\
L(I) & 33.97 & \textbf{47.69} & 20.64 & \textbf{21.12} & \textbf{38.39} & 161.81\\
L(I\^{}2) & 29.27 & 46.98 & 21.11 & \textbf{21.56} & 37.9 & 156.82\\
L(T) & \textbf{41.32} & 47.13 & \textbf{24.8} & \textbf{20.83} & 37.02 & \textbf{171.1}\\\hline
\end{tabu}
}
\caption{IPC, IPC\^{}2 and time score of the learned planning strategies and relevant baselines on 60 test problems randomly generated for each of the five domains: \emph{Transport} (T), \emph{Parking} (P), \emph{Elevators} (E), \emph{No-mystery} (N) and \emph{Floortile} (F). Average of 10 test runs. L(I), L(I\^{}2) and L(T) are the strategies trained with rewards based on IPC, IPC\^{}2 and time-based score, respectively. Values within 1 point from the highest one highlighted in \textbf{bold}.}
\label{tab:results}
\end{table}

\begin{table}[ht]
\centering
{\small \begin{tabu}{|c|ccccc|c|}
\hline
\multicolumn{7}{|c|}{IPC score}\\\hline
& T & P & E & N & F & Sum \\\hline
GBFS & 0 & 5.89 & \textbf{11.56} & \textbf{8} & 3.62 & 29.07\\
$\epsilon$-gr.& 0 & 3.84 & \textbf{11.68} & \textbf{7.72} & 4.87 & 28.11\\
RW & 0.92 & 4.55 & 9.88 & 6.91 & 2.6 & 24.86\\
Local & \textbf{2} & 9.98 & \textbf{11.63} & 6.98 & 3.64 & 34.23\\
DFS & 0 & 8.88 & 7.47 & \textbf{7.77} & 0 & 24.12\\
Uni & 0 & 6.21 & \textbf{10.81} & \textbf{7} & 3.61 & 27.63\\
L(I) & 0 & \textbf{12.27} & \textbf{11.78} & \textbf{7.72} & 5.7 & \textbf{37.47}\\
L(I\^{}2) & 0 & \textbf{12.09} & \textbf{11.15} & \textbf{7.72} & \textbf{6.72} & \textbf{37.68}\\
L(T) & 0.48 & 7.59 & 7.67 & \textbf{7.72} & 4.76 & 28.22\\\hline
\multicolumn{7}{|c|}{IPC\^{}2 score}\\\hline
& T & P & E & N & F & Sum \\\hline
GBFS & 0 & 5.79 & \textbf{11.15} & \textbf{8} & 3.29 & 28.23\\
$\epsilon$-gr.& 0 & 3.7 & \textbf{11.39} & \textbf{7.48} & 4.75 & 27.32\\
RW & 0.85 & 4.15 & 8.05 & 6.83 & 2.29 & 22.17\\
Local & \textbf{2} & 8.42 & 9.78 & 6.96 & 3.32 & 30.48\\
DFS & 0 & 7.38 & 4.63 & \textbf{7.55} & 0 & 19.56\\
Uni & 0 & 4.9 & 8.8 & \textbf{7} & 3.29 & 23.99\\
L(I) & 0 & \textbf{10.28} & 9.6 & \textbf{7.48} & \textbf{5.48} & \textbf{32.84}\\
L(I\^{}2) & 0 & \textbf{10.69} & 9.32 & \textbf{7.48} & \textbf{6.48} & \textbf{33.97}\\
L(T) & 0.23 & 6.5 & 4.82 & \textbf{7.48} & 4.55 & 23.58\\\hline
\multicolumn{7}{|c|}{Time score}\\\hline
& T & P & E & N & F & Sum \\\hline
GBFS & \textbf{0} & 4.03 & 10.35 & \textbf{7.77} & 3.99 & 26.14\\
$\epsilon$-gr.& \textbf{0} & 2.95 & 10.07 & \textbf{7.86} & 4.96 & 25.84\\
RW & \textbf{0.72} & 3.05 & 10.3 & 6.3 & 2.97 & 23.34\\
Local & \textbf{0.56} & 8.49 & \textbf{10.81} & \textbf{6.92} & 3.97 & 30.75\\
DFS & \textbf{0} & 8.47 & \textbf{11.67} & \textbf{7.32} & 0 & 27.46\\
Uni & \textbf{0} & 5.51 & \textbf{10.76} & 6.46 & 3.98 & 26.71\\
L(I) & \textbf{0} & \textbf{10.85} & 10.65 & \textbf{7.86} & \textbf{5.93} & \textbf{35.29}\\
L(I\^{}2) & \textbf{0} & \textbf{10.33} & \textbf{11.02} & \textbf{7.73} & \textbf{6.9} & \textbf{35.98}\\
L(T) & \textbf{0.13} & 7.28 & \textbf{11.64} & \textbf{7.89} & 4.95 & 31.89\\\hline
\end{tabu}}
\caption{IPC, IPC\^{}2 and time score of the learned planning strategies and relevant baselines on IPC-11 problems under 60 second time limit. \emph{Transport} (T), \emph{Parking} (P), \emph{Elevators} (E), \emph{No-mystery} (N) and \emph{Floortile} (F) domains. L(I), L(I\^{}2) and L(T) are the strategies trained with rewards based on IPC, IPC\^{}2 and time-based score, respectively. Values within 1 point from the highest one highlighted in \textbf{bold}.}
\label{tab:results-ipc-60}
\end{table}

The learning planner was implemented on the basis of the Fast Downward planning system \cite{Helmert2006}. The source code is available online\footnote{\url{https://github.com/pgomoluch/fd-learn}}.

We tested the system on five IPC domains of \emph{Transport}, \emph{Parking}, \emph{Elevators}, \emph{No-mystery} and \emph{Floortile}. This is the set of domains used in the learning track of IPC 2014 \cite{Vallati2015}, with the exception of the \emph{Spanner} domain. We excluded \emph{Spanner} because it was designed not to work well with delete-relaxation heuristics, such as the FF heuristic, used throughout all of our experiments. We fixed the time allocated for solving a single problem to 5 seconds. For each of the domains we generated 1000 training problems using the problem generators from the learning track of IPC 2014\footnote{\url{http://www.cs.colostate.edu/~ipc2014/}}. The parameters passed to the problem generators were selected to match the timeout of 5 seconds: for each of the domains we aimed at parameters for which the resulting problems will prove challenging but possible to solve. More precisely, we searched for generator configurations, for which about 50\% of the problems could be solved in 5 seconds by the baseline planner using GBFS guided by FF heuristic \cite{Hoffmann2001} with unary operator costs, which we used as the heuristic function throughout all the experiments. Training on substantially larger problems would be difficult in the current framework, because of the time required to complete a single episode (that is, solve the planning problem) and the sparser reward (more actions contributing to a single solution).

For every combination of domain and reward function, we trained the policy on a single CPU for 48 hours. During this time, problems were sampled randomly from the training set and attempted $N=5$ times, before the policy was updated according to Equation \ref{eq:update-rule} with $\alpha=0.02$. 
Table \ref{tab:policies} shows the policies learned for each of the domains and reward functions. Every entry of the tables indicates the probability of selecting the routine given by the column in the state given by the row.

\begin{figure*}
\centering
\begin{tabu}{cc}
\includegraphics[width=.95\columnwidth]{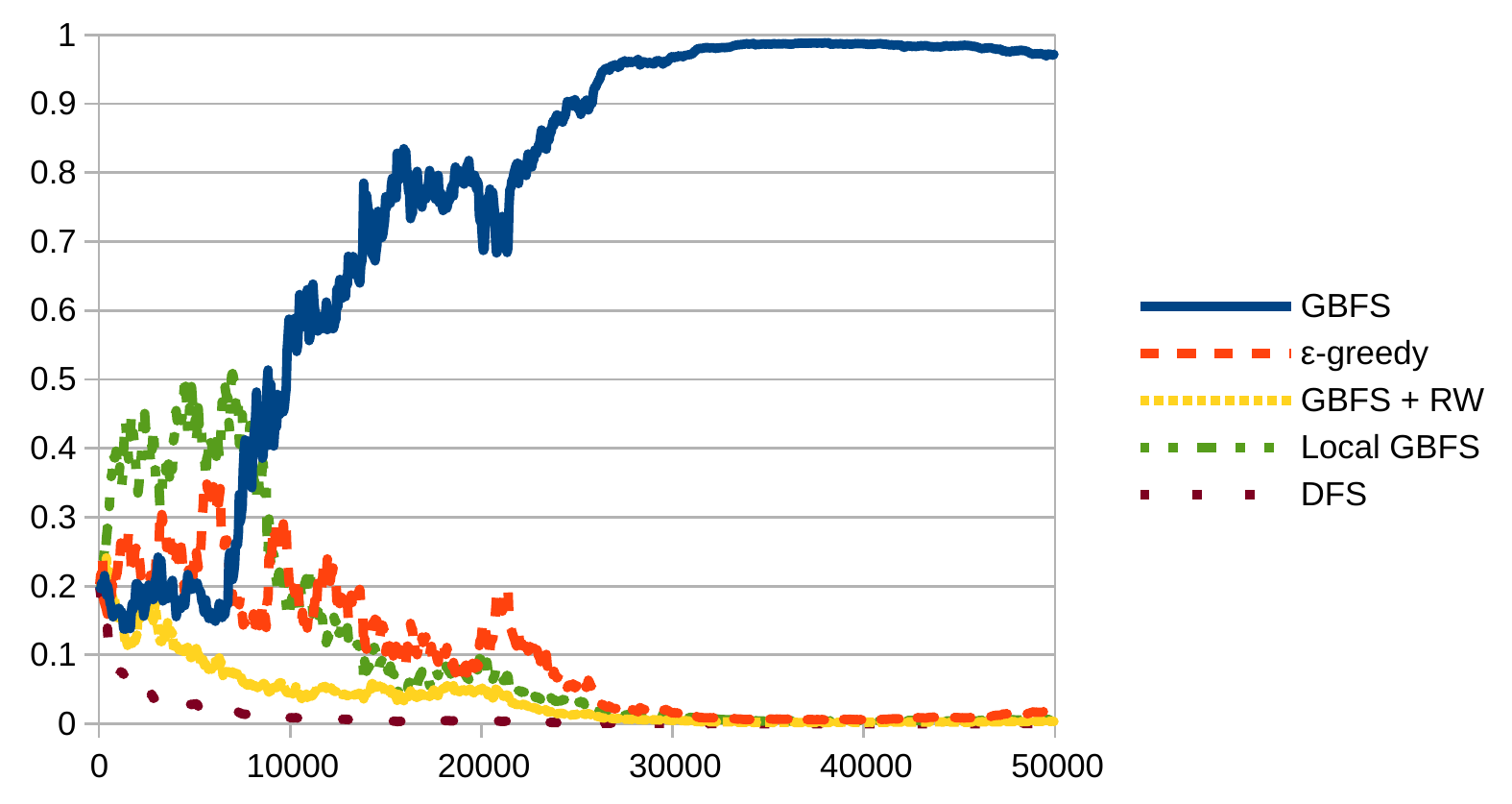}&
\includegraphics[width=.95\columnwidth]{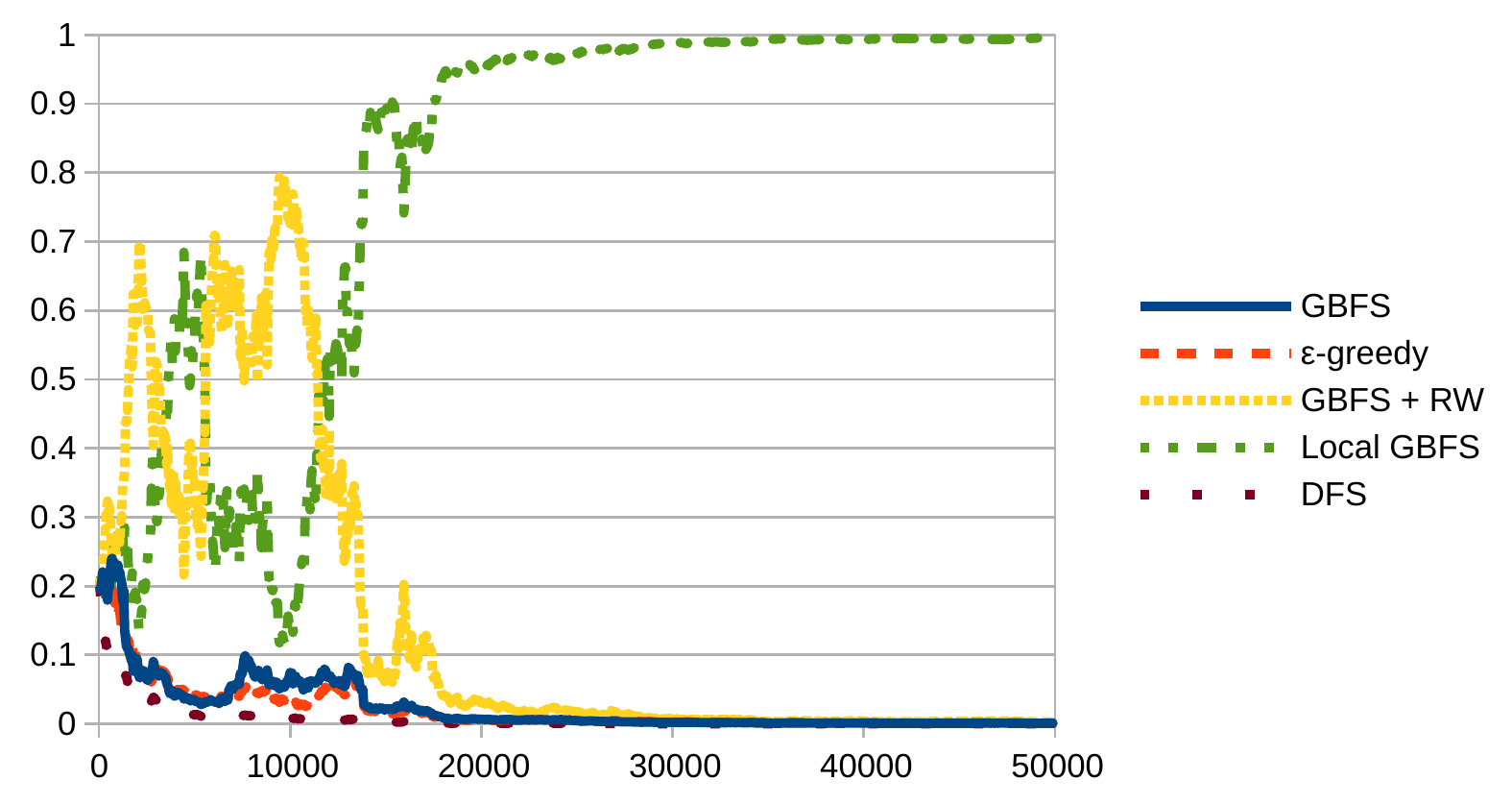}\\
\emph{Transport}, IPC reward, \emph{near early}& \emph{Transport}, IPC reward, \emph{near late}\\\\
\includegraphics[width=.95\columnwidth]{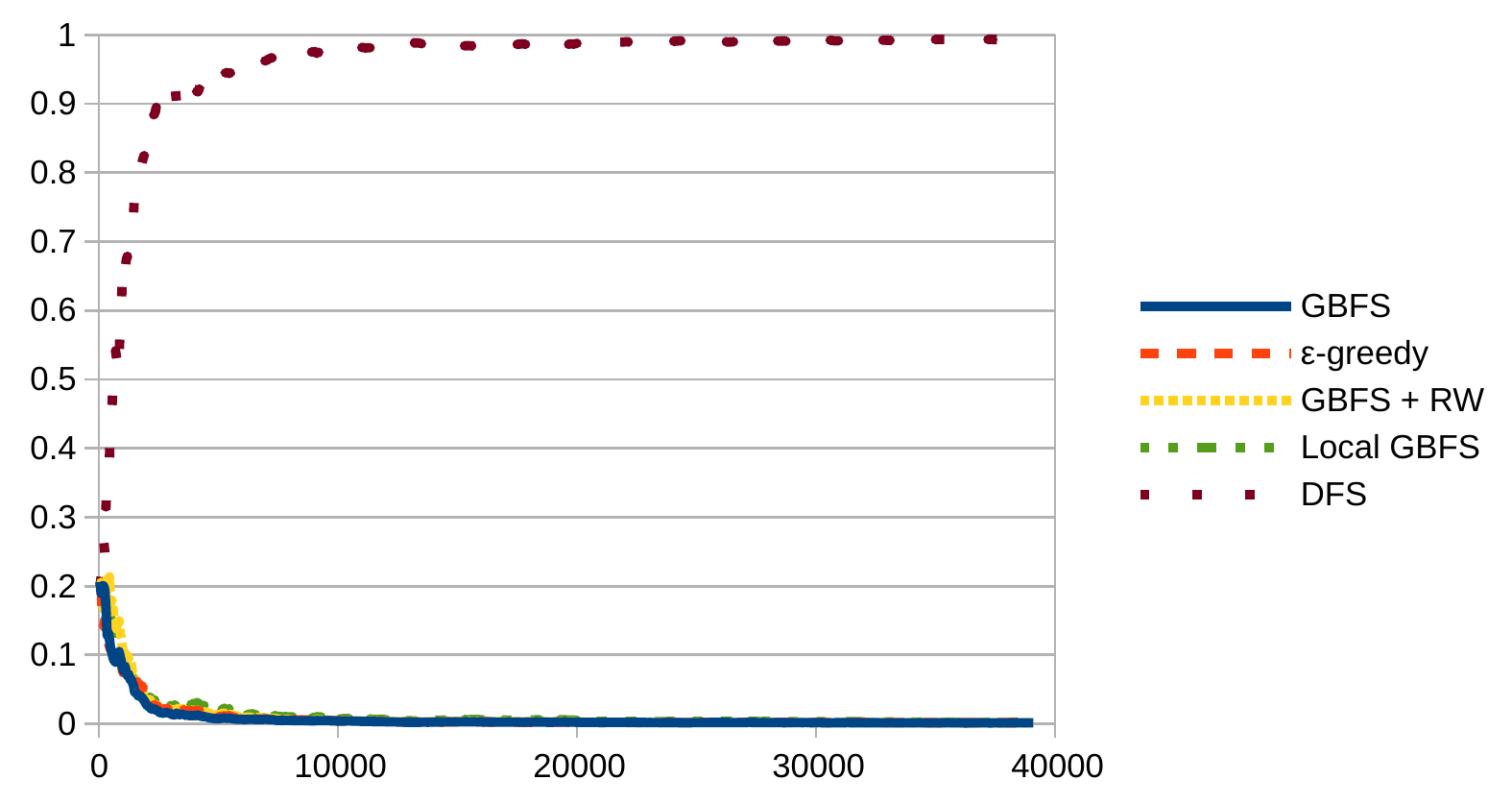}&
\includegraphics[width=.95\columnwidth]{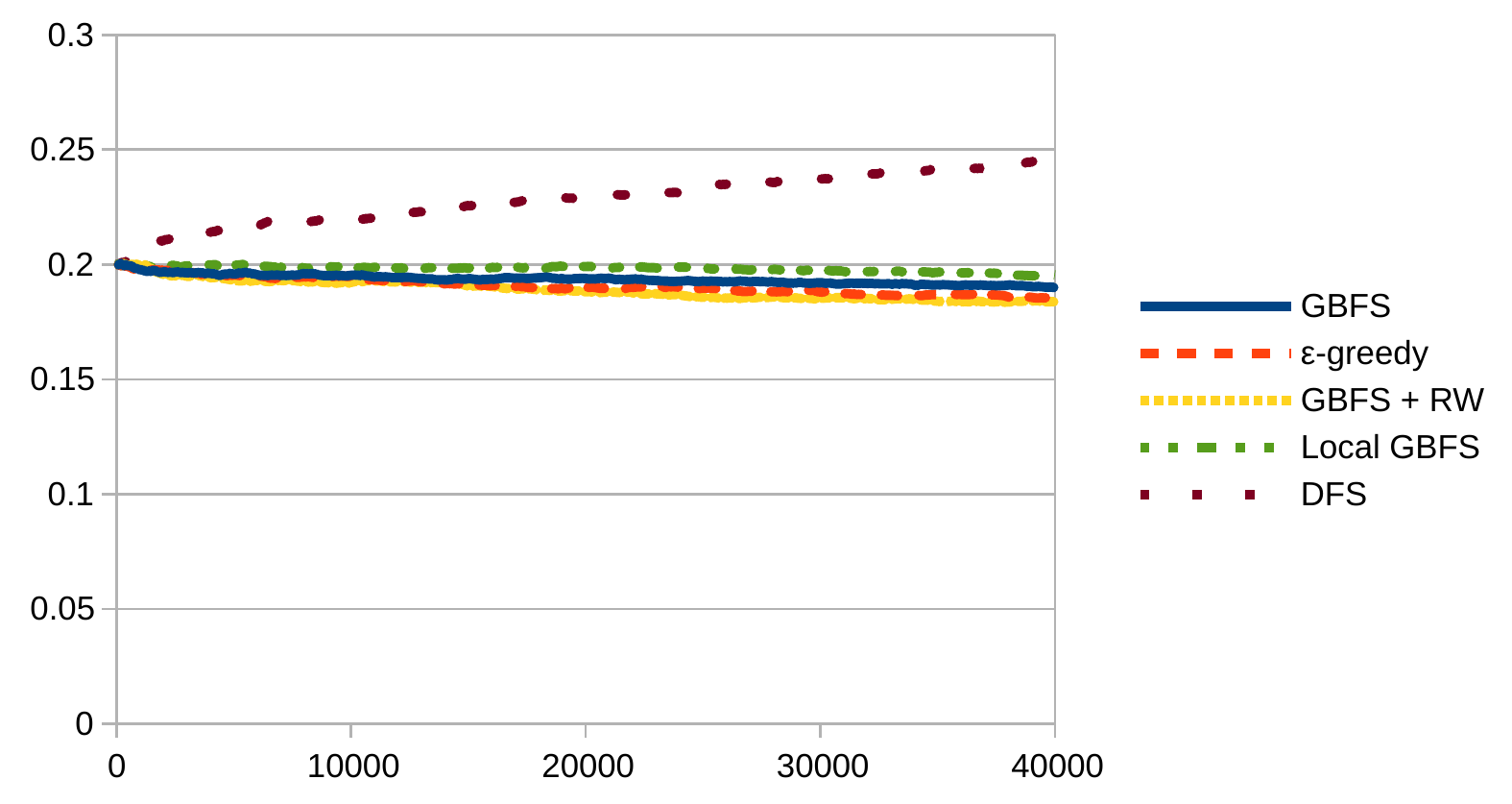}\\
\emph{Parking}, IPC reward, \emph{near early}& \emph{Parking}, time reward, \emph{near late}\\\\
\includegraphics[width=.95\columnwidth]{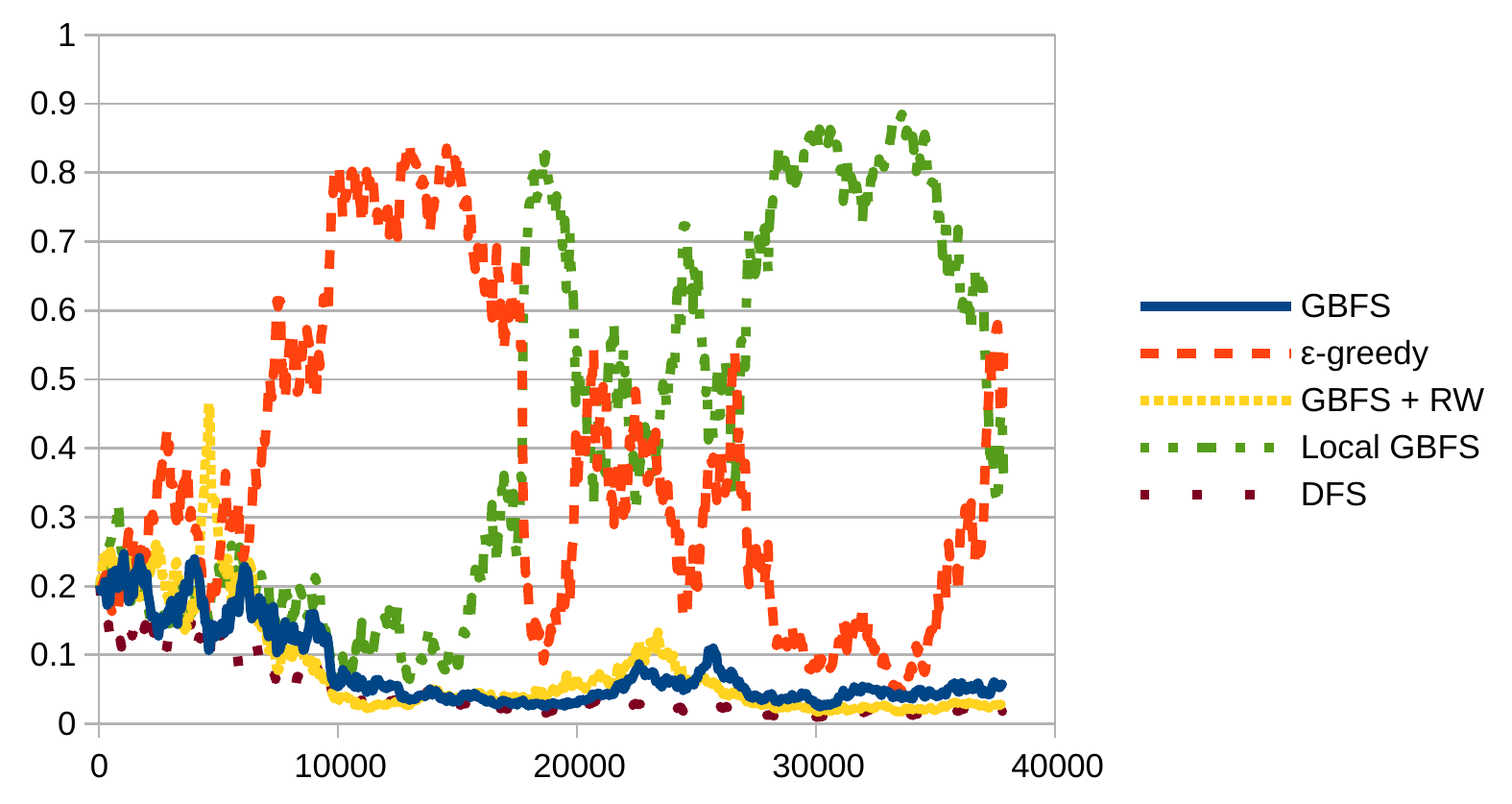}&
\includegraphics[width=.95\columnwidth]{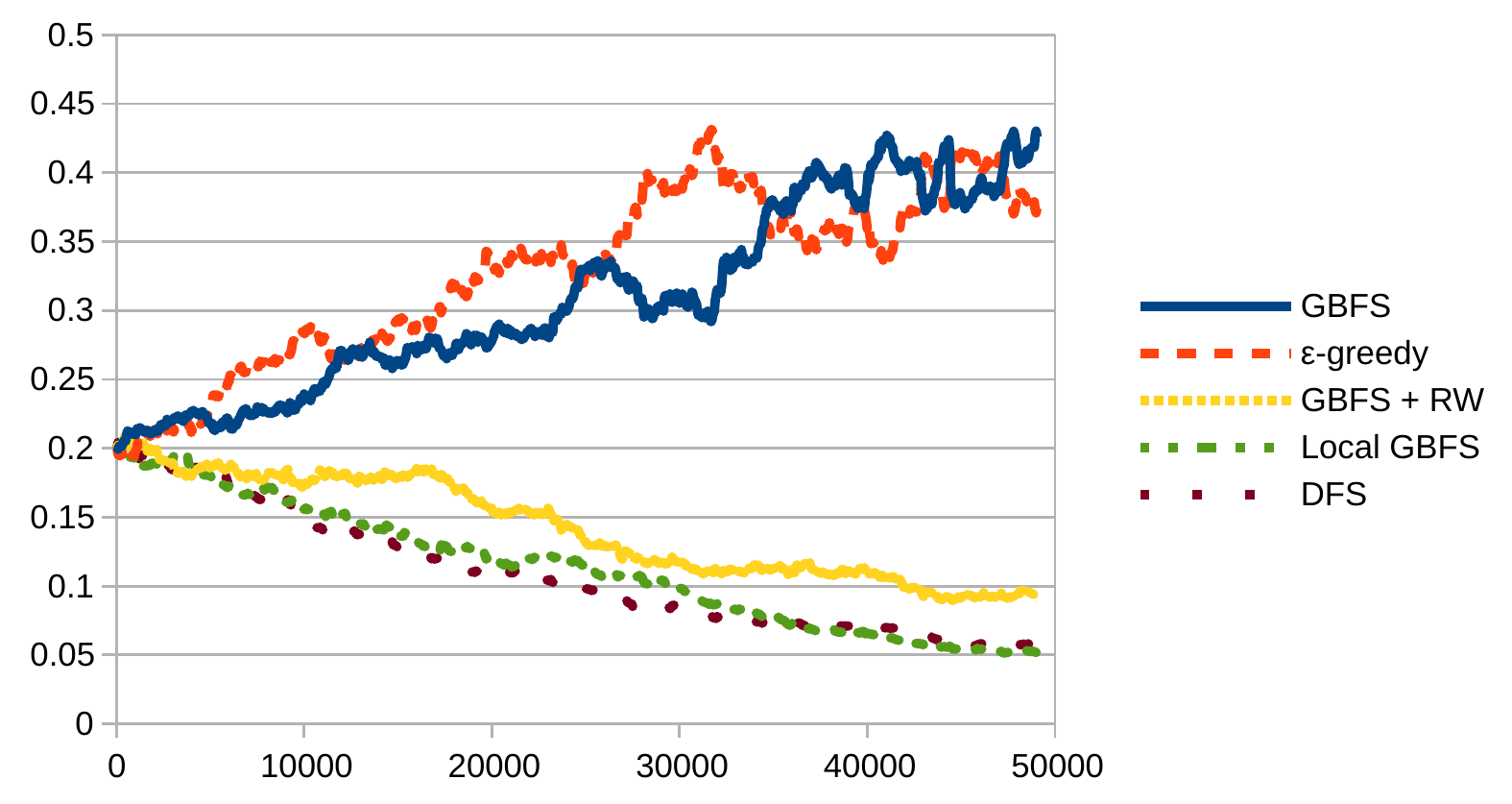}\\
\emph{Elevators}, IPC reward, \emph{near early}& \emph{No-mystery}, time reward, \emph{near late}\\\\
\end{tabu}
\caption{
Policies for selected states and reward functions, changing with the number of training episodes.
The policy for \emph{near early} state in \emph{Transport} domain under the IPC reward (top left) chooses GBFS nearly deterministically. DFS is discarded particularly early because in \emph{Transport} it leads to plans of very high cost.
For the \emph{near late} state (top right), the learner oscillates between greedy search with random walks and local search, before committing to the latter.
The policy for \emph{near early} state in \emph{Parking} domain under the IPC reward (middle left) quickly commits to DFS, which is typical for the \emph{Parking} domain.
However, for \emph{near late} state and the time reward (middle right), the policy remains close to uniform despite preference for DFS. This is because the state is rarely visited, providing few samples.
For \emph{near early} state in \emph{Elevators} domain under IPC reward (bottom left) the learner does not find a stable policy and oscillates between local and $\epsilon$-greedy search, which both seem reasonable choices.
For \emph{near late} state in \emph{No-mystery} domain under time reward (bottom right), the policy remains nondeterministic after 48h of training, choosing either of the preferred routines with similar probability. GBFS and $\epsilon$-greedy search are generally preferred over more aggressive routines in the \emph{No-mystery} domain, which was designed to enforce nearly-optimal solutions.
}
\label{fig:training}
\end{figure*}

In the majority of the states, the learned policy approaches a deterministic one, with the probability of choosing the dominant routine exceeding 0.9. For some states, however, the policy remains nondeterministic. This is often the case for states visited infrequently during the training process. The most extreme example is the \emph{far late} state, which did not occur at all when training on \emph{Parking} with the IPC or time reward. To shed some light on the learning process, in Figure \ref{fig:training} we plot the probabilities of choosing particular routines in selected states, as functions of the number of training episodes.

In the \emph{Transport} domain the preferred routine changes over the course of the search. For example, under the IPC reward, GBFS is preferred when substantial progress towards the goal has been made and there is still a lot of time (the \emph{near early} state). However, when the time starts to run out, the planner switches to BFS augmented with random walks (the \emph{near late} state). The learned policies vary depending on the selected reward function. The relatively conservative routines of GBFS and $\epsilon$-greedy search are chosen in two states under the IPC\^{}2 reward, one state under the plain IPC reward, and in none of the states in the case of time-based reward. In fact, with the time reward the learner goes as far as choosing DFS in the \emph{near early} state. Indeed, in this domain DFS allows for rapid progress towards the goal, at the cost of generating strongly suboptimal plans. The impact of changing the objective is even more visible in the \emph{Elevators} domain, where time reward results in frequent selection of DFS, while less aggressive options of local search and random walks are preferred under the two remaining rewards.

Since, in \emph{Parking}, the use of DFS comes without substantial increase of the plan costs, the learner chooses it very consistently, irrespective of the selected reward function. The opposite occurs for \emph{No-mystery} and \emph{Floortile}: since aggressive exploration can be harmful in these domains (because of dead ends), the learner confined itself to GBFS and $\epsilon$-greedy search. Again, this resulted in very similar policies learned across all three reward functions.

To evaluate the learned policies, for every domain we randomly generated 60 test problems, using the same generator parameters as for the training problems. Table \ref{tab:results} shows performance of the learned planners, measured by IPC, IPC\^{}2 and time-based score respectively. For comparison, the baselines of uniform random policy and each of the routines on its own are included. The strategies learned with the corresponding reward function consistently outperform most of the baselines, reaching the highest or nearly-highest scores. In particular, for \emph{Transport} the scores obtained when learning with relevant reward are higher than those reached with any of the routines on their own. In \emph{Elevators}, the learner scores a bit lower than the best single routine (local search for IPC, GBFS for IPC\^{}2 and DFS for the time score), but remains ahead of most others. On \emph{Parking}, \emph{No-mystery} and \emph{Floortile} all the strategies are comparable to the single best routine, which is expected given the policies learned in these domains.

A real-world application of a learning planner would aim at training the system on problems representative of the ones encountered in actual operation, e.g. by recording past problems. However, to check whether the learned policies can also be useful on problems larger than the training ones, we performed an additional series of experiments. We used the problem sets from IPC-2011, which is the last edition in which the five domains occurred together in the satisficing track. Initially, we conducted the experiments with the standard per-problem time limit of 30 minutes. The value of timeout $T$, used for computing the time score was adjusted accordingly to 1800s. However, the \emph{Parking} and \emph{Elevators} problems turned out to be very easy even for the plain GBFS(FF), which only failed one of 20 \emph{Parking} problems. GBFS with random walks also failed one of the problems, but otherwise all of the \emph{Parking} and \emph{Elevators} problems were solved by all of the planners. In this situation, the learned planners were unable to match GBFS, which generally returned solutions of lower cost. This is expected given the added depth-bias of the other routines and therefore the strategies relying on them.

For this reason, we decided to reduce the time limit to one minute (again, $T$ was adjusted to 60s). This setting was intended to reflect more closely the relative difficulty of the training setting, in which, as stated at the beginning of this section, only about half of the problems could be solved in time by GBFS with FF heuristic. Indeed, with one minute timeout GBFS solved 6 of the \emph{Parking} problems and 12 of the \emph{Elevators} problems. The scores are reported in Table \ref{tab:results-ipc-60}. The strategies trained with IPC and IPC\^{}2 rewards remain competitive, although their advantage over the baselines is not as large as on problems generated from the training distributions. The strategies learned with the time-based reward generalize worse, but on aggregate remain ahead of the baselines in terms of the obtained time score.

\section{Conclusion and future work}
\label{sec:conclusion}
In this paper, we introduced a planning framework capable of alternating between various search techniques while solving a problem. We modeled the problem of choosing the planning technique a reinforcement learning problem. Further, we provided an instantiation of the framework, defining a set of compatible search routines and a high-level representation of the planner's state. The resulting system was trained and evaluated on five planning domains and with three different performance measures. The experimental results show that the learned strategies obtain good performance on all five domains. Furthermore, despite being trained on relatively small problems, the strategies were also useful on larger problems, provided that the time interplay between problem size and time constraint roughly matched the training setting.

Our future work will investigate more complex representations of the planner's state and ways to improve practical sample efficiency of the learner. Another research direction is to extend the set of routines available to the learner. Possible additions include local search guided by other heuristic functions, width-based search \cite{Lipovetzky2012}, and stochastic rollouts driven by preferred operators.


\bibliographystyle{aaai}
\bibliography{references}

\end{document}